\documentclass[a4paper,12pt]{article}
\usepackage[paper=letterpaper,top=0.75in,bottom=0.75in,right=0.625in,left=0.625in]{geometry}% http://ctan.org/pkg/geometry
\usepackage{cite}
\usepackage{amsmath,amssymb,amsfonts}
\usepackage{algorithmic}
\usepackage{graphicx}
\usepackage{textcomp}
\usepackage{subcaption}
\usepackage{float}
\usepackage{array}
\usepackage{varwidth}
\usepackage[numbers]{natbib}
\usepackage{multirow}
\usepackage[table]{xcolor}
\usepackage{makecell}
\usepackage{verbatim}
\usepackage{caption}
\usepackage{authblk}
\usepackage{enumitem}
\graphicspath{{./}}

\providecommand{\keywords}[1]
{
  \small	
  \textbf{\textit{Keywords---}} #1
}

\date{\vspace{-5ex}}
\begin{document}
%\newgeometry{top=1in,bottom=0.75in,right=0.625in,left=0.625in}
%\newgeometry{top=1in,bottom=0.75in,right=0.63in,left=0.63in}

\title{\textbf{A Socially Adaptable Framework for Human-Robot Interaction}} %[A Socially Adaptable Framework for HRI]

\author[1,2,3]{Ana Tanevska}
\author[1]{Francesco Rea}
\author[1]{Giulio Sandini}
\author[3]{Lola Ca\~namero}
\author[4]{\\Alessandra Sciutti}
\affil[1]{Department of Robotics, Brain and Cognitive Science, Italian Institute of Technology (IIT), Genova, Italy}
\affil[2]{DIBRIS, University of Genova, Genova, Italy}
\affil[3]{EECAiA Lab, School of Computer Science, University of Hertfordshire, Hatfield, United Kingdom}
\affil[4]{Cognitive Architecture for Collaborative Technologies Unit, Italian Institute of Technology (IIT), Genova, Italy }

\maketitle

\begin{abstract}
In our everyday lives we are accustomed to partake in complex, personalized, adaptive interactions with our peers. For a social robot to be able to recreate this same kind of rich, human-like interaction, it should be aware of our needs and affective states and be capable of continuously adapting its behavior to them. One proposed solution to this problem would involve the robot to learn how to select the behaviors that would maximize the pleasantness of the interaction for its peers, guided by an internal motivation system that would provide autonomy to its decision-making process. 	

We are interested in studying how an adaptive robotic framework of this kind would function and personalize to different users. In addition we explore whether including the element of adaptability and personalization in a cognitive framework will bring any additional richness to the human-robot interaction (HRI), or if it will instead bring uncertainty and unpredictability that would not be accepted by the robot`s human peers.

To this end, we designed a socially-adaptive framework for the humanoid robot iCub. Thanks to it, the robot perceives and reuses the affective and interactive signals from the person as input for the adaptation based on internal social motivation. We strive to investigate the value of the generated adaptation in our framework in the context of HRI. In particular, we compare how users will experience interaction with an adaptive versus a non-adaptive social robot. 

To address these questions, we propose a comparative interaction study with iCub where users act as the robot's caretaker, and iCub's social adaptation is guided by an internal comfort level that varies with the amount of stimuli iCub receives from its caretaker. We investigate and compare how the internal dynamics of the robot would be perceived by people in a condition when the robot does not personalize its interaction, and in a condition where it is adaptive. Finally, we establish the potential benefits that an adaptive framework could bring to the context of having repeated interactions with a humanoid robot.

\end{abstract}


%%%%%%%%%%%%%%%%%%%%%%%%%%%%%%%%%%%%%%%%%
\section{Introduction}
People have a natural predisposition to interact in an adaptive manner with others, by instinctively changing their actions, tones and speech according to the perceived needs of their peers \cite{lindblom1990explaining}\cite{savidis2009unified}. Moreover, we are not only capable of registering the affective and cognitive state of our partners, but over a prolonged period of interaction we also learn which behaviors are the most appropriate and well-suited for each one of them individually \cite{mehrabian1972measure}. This universal trait that we share regardless of our different personalities is referred to as social adaptation (adaptability). Humans are very often capable of adapting to others, even though our personalities may influence the speed and efficacy of the adaptation. This means that in our everyday lives we are accustomed to partake in complex and personalized interactions with our peers. \\

Carrying this ability to personalize to HRI is highly desirable since it would provide user-personalized interaction, a crucial element in many HRI scenarios - interactions with older adults \cite{kidd2006sociable}\cite{broadbent2011human}\cite{sharkey2014robots}, assistive or rehabilitative robotics \cite{wood2017iterative}\cite{plaisant2000storytelling}\cite{admoni2014data}, child-robot interaction \cite{paiva2014emotion}\cite{tanaka2012children}, collaborative learning \cite{ramachandran2016shaping}\cite{jimenez2015emotional}, and many others. For a social robot to be able to recreate this same kind of rich, human-like interaction, it should be aware of our needs and affective states and be capable of continuously adapting its behavior to them \cite{ahmad2019robot}\cite{vaufreydaz2016starting}\cite{breazeal1999build}\cite{canamero2006attachment}\cite{kishi2014bipedal}.\\

Equipping a robot with these functionalities however is not a straightforward task. One potentially robust approach for solving this might consist of implementing a framework for the robot supporting social awareness and adaptation. In other words, the robot would need to be equipped with the basic cognitive functionalities, which would allow it to learn how to select the behaviors maximizing the pleasantness of the interaction for its peers, while being guided by an internal motivation system that would provide autonomy to its decision-making process.\\

In this direction, the goal of our research was threefold: attempt to design a cognitive architecture supporting social HRI and implement it on a robotic platform; study how an adaptive framework of this kind would function when tested in HRI studies with users; and explore how including the element of adaptability and personalization in a cognitive framework would in reality affect the users - would it bring an additional richness to the human-robot interaction as hypothesized, or would it instead only add uncertainty and unpredictability that would not be accepted by the robot`s human peers?\\

In our past works, we have explored adaptation in child-robot interaction (CRI) in the context of switching between different game-based behaviors \cite{tanevska2016evaluation}. The architecture was affect-based \cite{tanevska2018designing}, and the robot could express three basic emotions (a "happy", a "sad", and a "neutral" state) in a simple way. These emotions were affected by the level of engagement the child felt towards the current robot's behavior. The robot aimed to keep the child entertained for longer by learning how the child reacted to the switch between different game modalities. We have since expanded on the core concept of a robot's internal state guiding the adaptation, and we advanced from the discrete emotional states and one-dimensional adaptation to a more robust framework. Starting from the work of Hiolle and Ca\~namero \cite{hiolle2012eliciting}\cite{hiolle2014arousal} on affective adaptability, we have modified our architecture to utilize as motivation the level of comfort of the robot, which is increasing when the robot is interacting with a person, and decreasing when it is left on its own.\\

The robotic platform selected for our study was the humanoid robot iCub \cite{metta2008icub}, and the scenario for testing the framework's functionalities was inspired by a typical interaction between a toddler and its caregiver, where the toddlers tend to seek the attention of their caretakers after being alone for a while, but as soon as their social need has been saturated they lose interest and turn their attention to something else \cite{feldman2003infant}. The robot therefore acted as a young child, asking the caretaker's company or playing on its own and the human partners could establish and maintain the interaction by touching the robot, showing their face and smiling, or showing toys to the robot. This scenario was deemed suitable for studying some fundamental aspects of interaction (such as initiation and withdrawal) with a fully autonomous robot behavior and very limited constraints to the human activities, as well as in a seemingly naturalistic context. Furthermore, we verified these assumptions over the course of several validation and pilot studies \cite{tanevska2019}\cite{tanevska2019smc}.\\

In this paper we cover the work we did on developing a cognitive framework for human-robot interaction; we analyze the various challenges encountered during the implementing of the cognitive functionalities and porting the framework on a robotic platform; and finally we present the user studies performed with the iCub robot, focused on understanding how a cognitive framework behaves in a free-form HRI context and if humans can be aware and appreciate the adaptivity of the robot. The rest of the paper is organized as follows: Section \ref{sect:architecture} presents the adaptive framework for our architecture, followed by Section \ref{sect:experimental} which presents the experimental methods applied in our study with iCub. Finally, in Sections \ref{sect:results} and \ref{sect:discussion} we present the findings from our study and we touch on our plans for future work.

%\restoregeometry

%%%%%%%%%%%%%%%%%%%%%%%%%%%%%%%%%%%%%%%%%
\section{Architecture}
\label{sect:architecture}
A cognitive agent (be it a natural or an artificial one) should be capable of \textit{autonomously} predicting the future, by relying on \textit{memories} of the past, \textit{perceptions} of the present, and \textit{anticipation} of both the behavior of the world around it as well as of its own actions  \cite{vernon2014artificial}. Additionally, the cognitive agent needs to allow for the uncertainty of its predictions and \textit{learn} by observing what actually happens after an action, and then assimilating that perceptive input into its knowledge about the world, \textit{adapting} on the way its behavior and manner of doing things.\\

Following this, cognition can be defined as the process by which an autonomous agent perceives its environment, learns from experience, anticipates the outcome of events, acts to pursue goals and adapts to changing circumstances. Our work focused on developing a cognitive architecture for autonomous behavior, supporting all of these functionalities, for generalized applicability on any robotic platform for HRI. \\

It is important to note that there are several well-studied cognitive architectures in literature which are designed for a more context-free human-robot interaction, or even in a broader sense for general intelligence, such as the ACT-R/E\cite{trafton2013act}, SOAR\cite{laird1987soar} and Sigma\cite{rosenbloom2016sigma}, to name just a couple; however we opted for a simpler approach in order to have more freedom for future expansions of the architecture. Our framework over its various developments was tested on the iCub humanoid robot. The architecture relies on the robot evaluating the affective state of its human peers and their mode of interacting with the robot as factors which determine the robot's own internal emotional condition, and subsequent choice of behavior.\\

Starting from this foundation, our framework for the iCub consisted of the following modules and their functionalities:
\begin{itemize}
\item Perception module, processing tactile and visual stimuli;
\item Action module, tasked with moving iCub's joint groups.;
\item Adaptation module, active only in the adaptive profile for the robot and in charge of regulating iCub's social need.
\end{itemize}

\subsection{Perception module}
The perception module was tasked with processing stimuli from two sensor groups: \emph{tactile stimuli} - the data processed from the skin sensor patches on the iCub on its arms and torso, which carried information about the size of the area that was touched (expressed in number of \emph{taxels} - tactile elements) and the average pressure of the touch\cite{cannata2008embedded}; and \emph{visual stimuli} - the images coming from iCub's eye camera, jointly analyzed for detecting the presence of a face and extracting the facial expression of the person, as well as for detecting the presence of some of iCub's toys. The module was realized using iCub's middleware libraries \cite{metta2008icub} for processing the data from the skin covers on its torso and arms; as well as using the open-source library OpenFace \cite{baltruvsaitis2016openface} for extracting and analyzing the facial features of the caretaker, represented by their facial action units (AUs) \cite{ekman1978facial}.\\

The data from the OpenFace library were analyzed for obtaining the most salient action units from the detected facial features. We considered as possitive-associated AUs smiling and cheek raising with crinkling of the eyes, and as negative-associated AUs brow lowering, nose wrinkling and raising the upper lip in a snarl. Presence of all positive AUs was classified as "smiling" (presence of just a mouth smile but no positive upper AUs signified a fake smile, and was not classified as "smiling"), presence of only the brow lowering but without additional negative AUs was classified as "contemplating" whereas the presence of all negative AUs signified "frowning". If neither of these AUs groups were present in the frame, the user's affective expression was classified as "neutral".

\begin{figure}[h]
    \centering
        \includegraphics[width=0.5\textwidth]{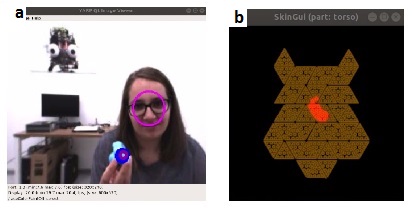}
    \caption{The two outputs from the perception module. (Informed consent of participants has been obtained for the use of their photo).}
    \label{fig:percInter}
\end{figure}

In addition to the affect detection functionality, the visual perception consisted also of the color detection functionality, which was able to detect and track a set of predefined colors, looking for contours in the image of a certain size (fitting the size of the toys) and color. Figure \ref{fig:percInter}.a shows the simultaneous detection and tracking of the face of the participant and a toy - the center of the face is indicated with a pink circle, the center of the object with a blue one, and the smaller purple circle instead indicates where iCub's attention is at that moment, i.e. which stimuli is tracking.  Figure \ref{fig:percInter}.b instead shows detected touch on the tactile covers of iCub's torso.  For the skin there was some additional processing post-extraction; as during prolonged interaction the tactile sensors tended to overheat and give phantom signals, the data were filtered to register as touch only areas that were larger than 5 taxels and recorded avg. pressure larger than 12.0. These data were processed for the torso and both arms separately, and sent to the perception module.

\subsection{Action module}
The action module communicated with iCub's middleware \cite{parmiggiani2012design} and performed a finite set of actions by controlling the specific body part in the joint space. iCub was holding a box of its toys and it could \emph{move its arms} in an extending or flexing motion, thus bringing the box closer to the person or away from them. A \emph{looking} action was performed every time iCub was changing its gaze focus, utilizing motions of the neck and saccadic motions of the eyes. When iCub wanted to engage with the caretaker, it would \emph{straighten up} and look for the person, and then during the interaction engage in gaze-cueing and looking at objects, whereas when iCub was oversaturated and wanted to disengage, it would \emph{lean down} to the table and away from the person, and look down to its toys, ignoring other attempts to engage.

\subsection{Adaptation module}
This module maintained iCub's comfort and guided the adaptation process. The motivation in our architecture was represented by iCub's striving to remain in an optimal level of comfort, which was achieved by continuously \emph{adapting} and changing the parameters of the motivation functionality. The comfort of iCub grew when a person was interacting with it, and the stimuli were weighted accordingly - a multimodal interaction (receiving both visual and tactile stimuli) or a longer, steadier interaction rated higher and increased the comfort faster. Inversely, lack of any stimuli caused the comfort value to decay. iCub's social architecture was also equipped with a saturation and a critical threshold, which were reached when the interaction was getting too intense or was too sparse, respectively.\\

At the beginning of the interaction with each user, iCub started with its comfort set at 50\% of the maximum value it could have. Then the comfort level was updated continuously at the beginning of each cycle of the control loop of the interaction\footnote{\label{loop}Referring here to the perception-action control loop of iCub's architecture}. This happened in the following manner:
\texttt{
\begin{tabbing}
if \= (F[t] || T[t]):  \\
\> C[t] = (F[t]+T[t]+C[t-1]$\tau$)/($\tau$+0.1)\\
else: \\
\> C[t] = $\beta$*C[t-1]
\end{tabbing}
}
where \emph{C[t]} indicates the current comfort level whereas \emph{C[t-1]} is the previous comfort level; \emph{F[t]} and \emph{T[t]} are the input stimuli from the visual and tactile sensors respectively. $\beta$ and $\tau$ are the social variables dictating the decay and growth rate of the comfort value, where their initial values were set at $\beta$ = 0.998 and $\tau$  = 500.\\

When there was a human interacting with the robot (iCub was perceiving a face in front of it, or registering touch with its skin), the comfort \emph{C[t]} at time \emph{t} was updated using the first formula, which takes into consideration both modalities in which the user could interact with iCub, as well as the previous level of comfort \emph{C[t-1]}; on the other hand if iCub was not currently engaged in interaction, its comfort was updated as depicted in the second formula, which calculated the decay of the comfort. \\

The variables $\tau$ and $\beta$ were the growth and decay rates respectively, which were part of the internal variables that iCub could modify in its adaptation process. $\tau$ modulated how much \emph{C[t-1]} was taken into consideration: a smaller $\tau$ bringing a more rapid growth of the comfort when stimuli were detected, and a larger value a slower, steadier growth. $\beta$ was indicating how quickly \emph{C[t]} decayed without stimuli; the smaller the value of $\beta$, the more drastic the decay of the comfort.\footnote{\label{explain} The manner of modifying the $\tau$ and $\beta$ variables was carried over from the related research done in \cite{hiolle2012eliciting}\cite{hiolle2014arousal}. A previous simulation study \cite{tanevska2019smc} explored more in depth how the behavior of the architecture could be affected by varying the initial values of these rates, using different steps in the adaptation, and starting with different critical and saturation thresholds.} \\

iCub's architecture allowed for adaptation on two dimensions - the frequency of interaction initiation and the duration of the interaction. The first one affected $\beta$, and the adaptation on the second dimension instead modulated $\tau$. After each instance of iCub adapting on either dimension, it entered a suspension period of 20 seconds where it attempted to recover and during which it was not open to interaction with the users. The adaptation process had the following pattern:
\begin{itemize}
\item If the comfort reached the saturation limit: increase the value of $\tau$  by 500, and during the period of suspension ignore all stimuli. The resulting lack of sensitivity to stimulation leads to a decrease in the comfort value back to the optimal zone.
\item If the comfort dropped to the critical level: increase the value of $\beta$ by 0.005, attempt to engage the caretaker; if ignored enter the suspension period and simulate stimuli to itself so as to recover back to the optimal comfort level.
\end{itemize}

To give a reference of the framework's dynamics - the initial values of the comfort variable, $\beta$, and $\tau$ provided for 1.5 minute of extreme interaction before hitting a threshold (1.5 minute of zero stimuli for a critical threshold, and 1.5 minute of full multimodal interaction for saturation). The time limits increased after each architecture adaptation, e.g. after 2 adaptations prompted by critical triggers, iCub could be left by itself for 7.5 minutes before hitting the threshold again. \\

Originally the architecture adapted by immediately resetting the comfort level back to the optimal level and continuing with the interaction. The suspension period was included as a factor only after the validation of the original architecture with participants, during which it was realized that a continuation of responsiveness of the robot might not have allowed for the participants to infer that they were doing something not ideal for the robot. For example -  in the case of saturation, after the instantaneous robot withdrawal, it was immediately ready again to respond, which induced participants again to continue to interact in the same manner and trigger again saturation.

%%%%%%%%%%%%%%%%%%%%%%%%%%%%%%%%%%%%%%%%%
\section{Experimental methods}
\label{sect:experimental}

We had already established in a previous exploratory study that a game-based interaction scenario would not provide the desired amount of affective expresiveness in participants \cite{tanevska2018hamburg}. Since we had also seen the effectiveness of a caretaking scenario in prior pilot and validation studies with the iCub robot \cite{tanevska2019}, we decided to continue in the same direction and expand the existing experimental setup. As before, the interaction scenario placed iCub in the role of a toddler exploring and playing with its toys, while the participants were tasked as the iCub's caretaker. 

\subsection{Participants}
Twenty-six participants in total took part of the caretaker study. The youngest participant was aged 18 and the eldest 58, with the average age being 32.6 years (SD = 11.98). The gender ratio between the participants was 15:10:1 (M:F:NBGQ\footnote{NBGQ - non-binary/genderqueer}).

\subsection{Experimental setup}
Since we had already explored the preference of participants for an adaptive dynamic robotic profile over a static scripted one \cite{tanevska2019}, we now placed the focus on a different task - evaluating in greater detail the effect of the adaptation modality in two otherwise equally dynamic and responsive behavior profiles. In that direction, the two different "personalities" of iCub were both equipped with the full cognitive architecture described in the previous section, with the only difference being that one profile had the adaptation functionality disabled.\\

In both behavior profiles iCub's behavior was guided by its social skills, and in both conditions iCub began the interaction with the optimal values of the growth and decay variables as selected after the simulation study \cite{tanevska2019smc}. The only variation in the profiles was that in the \emph{fixed} profile (F) the values remained unchanged throughout the interaction (regardless of how many times the boundaries were hit), whereas in the \emph{adaptive} profile (A) instead there was the personalization of the architecture to each participant by modifying the values after each threshold hit.\\

The interaction between iCub and the participants was mostly free-form; and while iCub could try during the session to also initiate interaction, or would actively ask for it after hitting a critical or saturation point; for the most part participants had the liberty of guiding the interaction. During the entire interaction iCub could receive and process stimuli from the participants which could be tactile (contact with the skin patches on iCub's arms and torso) and visual (either observing the participant's face at an interacting distance and evaluaing the facial expressions, or detecting toys by recognizing their color and shape).\\

\begin{figure}[H]
    \centering
        \includegraphics[width=0.5\textwidth]{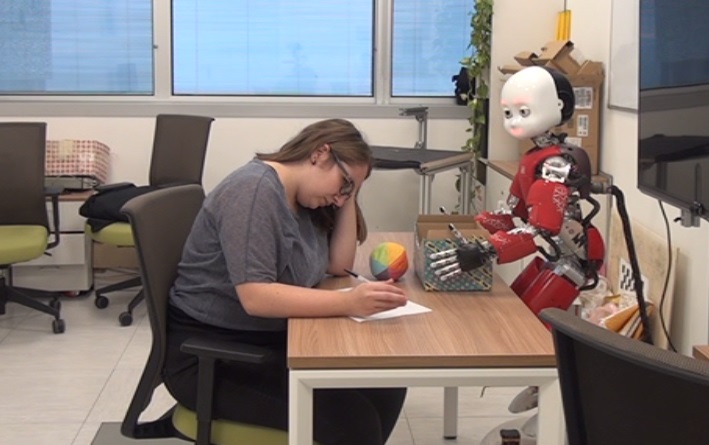}
    \caption{The layout of the laboratory setup. Informed consent of participants has been obtained for the use of their photo.}
    \label{fig:icublab1}
\end{figure}

In the laboratory iCub was positioned in front of a table (as shown in Figure \ref{fig:icublab1}), holding a box with toys, some of which were out of the box and spread across the table at the beginning of the interaction. The participants were offered a chair in front of the table facing iCub, but they also had the freedom to sit or walk anywhere in the room. 

When iCub was in a state of interacting with its caretaker, it maintained mutual gaze and tracked the person's face, or if the person was playing around with some of the toys it would track the toy that was nearest to it. If the person was not showing any toys to iCub, it would occassionally break mutual gaze and try to indicate toys to the person by looking down at a toy and back to the person (gaze-cueing), by saying the name of the toy or by moving the box towards the participant. In order to avoid giving participants the impression that iCub could understand them, the verbal utterings (which were the names of the colours iCub could recognize, as well as some encouraging and protesting sounds in order to attract attention or to disengage) were recorded in Macedonian and then processed and low-pass filtered so as to both make them sound more robotic as well as unintelligible to participants. %REWRITE IN MORE TECHNICAL WAY SIMILAR TO ACTION MODULE %%%

\subsection{Secondary task}
With the goal of further exploring the potential benefits of having critical and saturation thresholds in the architecture, we devised and approach to manipulate the behavior of the participants by introducing a timed secondary task at a certain point in the interaction. While in the pilot study any threshold hits were due to the behavior of the participants themselves and their way of interacting, there was not a possibility to observe what would the behavior look like if participants suddenly had a secondary task they needed to fulfill but the robot was still asking for their attention.\\

For this, a task needed to be considered that would involve a cognitive load on the participants, while at the same time being a task that would neither be too time-consuming (like sudoku), nor too attention-demanding or distracting (like a phone call during which participants would be tasked to write down some information). The solution selected was to present participants with some easier mathematical problems involving the basic arithmetic operations, which meant finding a set of numeric puzzles that would be both simple enough to do in a short time interval, but also appealing and interesting. The final choice for the secondary task was the pollinator puzzle\footnote{https://mathpickle.com/project/pollinator-puzzles/}.\\

\begin{figure}[H]
    \centering
        \includegraphics[width=0.5\textwidth]{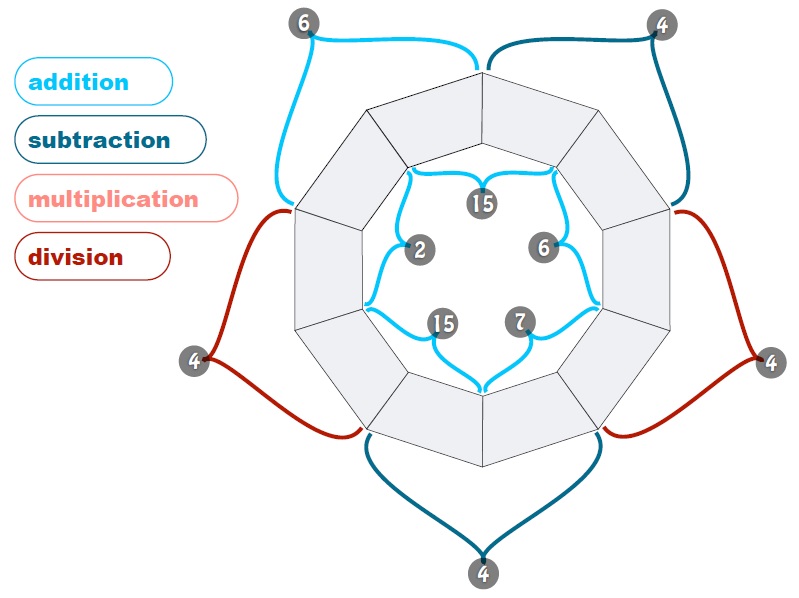}
    \caption{Sample pollinator puzzle}\label{fig:poll}
\end{figure}

The pollinator puzzle is a logic-based, combinatorial number-placement puzzle, where ten empty fields are arranged in a flower-like shape (see Figure \ref{fig:poll}). The digits 0-9 need to be placed in the empty fields, each digit appearing one time only without repetitions, in such a way that each pair of digits gives the specified result for the operation on the petals. Each puzzle has only one possible solution following these rules.

\subsection{Protocol}
Participants were evenly distributed in two groups of 13 people, where one group interacted first in the adaptive and then in the fixed dynamic setting, and the other vice versa. A session of interaction in either profile setting lasted 12 minutes, divided in three intervals of 4 minutes, the middle one of which was the interval when participants were asked to work on the secondary task.\\

Between the two sessions of interaction, as well as at the beginning and end of the interaction participants answered questionnaires (more details on the questionnaires in the following subsection), bringing the total time of commitment for the participants at around 45-50 mins. There were two environments in which the participants were stationed during their visit - the office setting and the laboratory setting.\\

Upon their arrival to the institute, participants were first brought to the office setting, where they were presented with the consent form and given time to read through it and sign it. Then, while still in the office, participants were informed that during the experiment there would be several moments during which they would be given different forms of questionnaires - related to their personality, relationship to iCub, as well as creativity and problem solving. This was followed by the familiarization phase for the pollinator puzzle. The concept and rules of the puzzle were explained to the participants, and they were presented with the first pollinator puzzle (the purpose of which was to obtain the baseline for each participant's performance). The participants were timed for 4 minutes (the amount of time allotted for the puzzle during the familiarization phase was the same as the time during the robot interaction). After the time ran out (or if participants completed the puzzle in less time - after they were done), we escorted the participants from the office and took them to the laboratory.\\

On the way to the laboratory we briefed participants on the experiment, more specifically they were told that they would have roughly half an hour of free interaction with the robot iCub Reddy, who is equipped with a toddler-like personality. We informed them of the modalities they could interact on with iCub, albeit in an informal way - "iCub can see you, it\footnote{the participants who spoke only Italian were briefed in Italian instead of English. Due to Italian not having a gender-neutral pronoun, iCub was referred to with "him" in Italian (lui)} can feel you when you pet it, it likes hearing you talk to it even though it does not understand you, it speaks its own language". Participants were purposefully informed that iCub likes hearing them because it was observed in our previous pilot study that people who knew iCub was not capable of speech recognition did not talk at all to the robot during the study.\\

Additionally participants were reassured that any perceived lack of interest or reciprocity on iCub's part would due to the robot switching its attention to something else (in line with its toddler personality), and not due to them interacting "in a wrong way". This was also deemed necessary to be included in the protocol due to a similar realization from the previous pilot that some people were getting worried when iCub would switch its attention and they thought they "did something wrong".

\subsection{Data Analysis}
The data collected during the study consisted of four main sources - the data collected from the questionnaires filled by the users; the evaluations from the filled pollinator puzzles; the video and audio recordings from the external camera; and the data collected by the robot during the interaction phases from the tactile sensors, internal camera and state machine output.

\subsubsection{Questionnaires}
Participants responded to questionnaires at three points during the interaction study - the first set of questionnaires was done after they entered the lab with the robot but before beginning with the interaction, the second set was halfway through the interaction (which in reality was the moment after which the robot switched personalities, unbeknownst to the participants), and the last set was at the end of the interaction.\footnote{these 3 points in the interaction are labelled as PRE, BETWEEN and POST}\\

All three sets of questionnaires collected the IOS rating of closeness between themselves and the robot \cite{aron1992inclusion}, as well as the Godspeed questionnaires on animacy and likeability \cite{bartneck2009measurement}. Additionally, in the second and third set of questionnaires there was also a qualitative question asking participants to describe the interaction using three adjectives, as well as a set of questions related to how they perceived the interaction with the robot. Finally in the third set of questionnaires there were two descriptive questions related to the different sessions, and the TIPI questionnaire. 

\subsubsection{Pollinator puzzle}
Participants did in total three rounds of the pollinator puzzle - one as a baseline before starting their interaction with the robot, one during the first interaction session and one during the second interaction session. There were two evaluation metrics for the puzzles - the \% of filled fields (out of the 10 empty fields) and the \% of accurately filled fields. \\

A combination metric was then designed in order to obtain a single evaluation value, where if X was the percentage of completeness and Y the percentage of accuracy, the final metric Z was obtained as Z = 0.4*X + 0.6*Y. The combination metric was designed with the goal of taking into account as factors both the accuracy and the completeness, but give a higher reward for the accuracy. 

\subsubsection{Internal data from iCub}
From the iCub itself we recorded the tactile and visual data, as well as all of the values of the architecture - the fluctuations of the comfort value and the changes to the decay and growth rates. The data from the architecture was annotated for each frame received by the robot with a timestamp and the state (of the state machine) iCub was in.
%%%%%%%%%%%%%%%%%%%%%%%%%%%%%%%%%%%%%%%%%
\section{Results}
\label{sect:results}
In this study we were interested in exploring whether adaptation is a necessary functionality for human-robot interaction, and particularly for the context of free-form social human-robot interaction? If there was not a clear task for the human to perform with the robot, would the adaptive functionality bring anything additional to the interaction? To address this, three related questions were formulated:
\begin{itemize} %put main
\item How much would the adaptive architecture change for each participant during the interaction, and how would people react to such personalization (answered in subsection \ref{archDyn})?
\item What would be the subjective evaluation of the participants for the interaction, and would it depend on the adaptivity of the robot? (answered in subsection \ref{subjEval})?
\item Would participants change their way of interaction across modalities or robot adaptivity level? (answered in subsection \ref{behavEval})?
\end{itemize}

\subsection{Architecture dynamics} 
\label{archDyn}
The cognitive framework developed for iCub was a continuously-changing one, learning by way of modifying its social variables and adapting to the person's frequency and intensity of interaction. This means that interacting with robot provoked changes in the internal states of iCub and its comfort level. Every time a threshold of the robot was hit, iCub adapted the appropriate comfort variable and its behavior changed accordingly.\\

If the critical threshold was hit, signifying lack of stable interaction with the person, iCub modified its decay rate and as a result could remain in an idle state for longer periods of time before it would need again to interact with the person. On the other hand, hitting the saturation threshold meant iCub was engaged with a person who was more intense in the way it behaved and interacted with iCub (using multiple modalities and interacting for a long stable period of time), so iCub modified its growth rate which enabled it to stay interacting for longer time.\\

\begin{figure*}
    \centering
        \includegraphics[width=\textwidth]{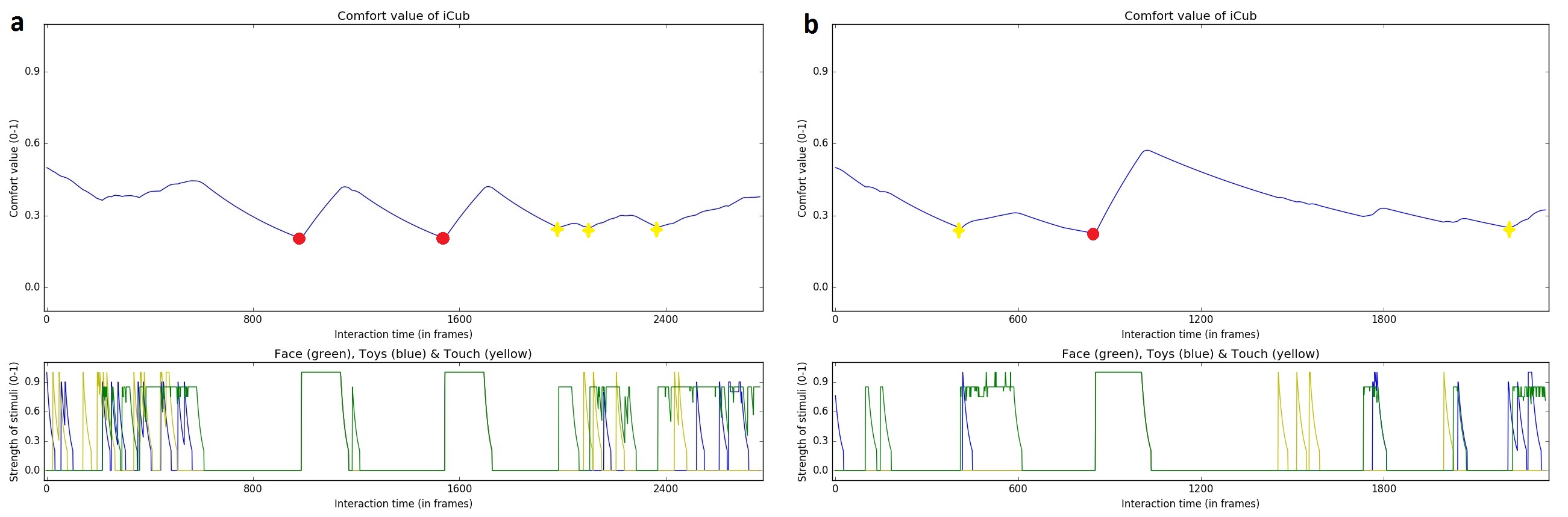}
    \caption{Architecture dynamics: upper graphs depict the variations in iCub's comfort value over the course of an interaction session, lower graphs depict the occurrence of stimuli. Critical hits where participants responded the robot's call for engagement shown in yellow stars, ignored critical hits shown in red dots. (a) FA participant interacting in F session, 5 points of hitting critical threshold. (b) AF participant interacting in A session, 3 points of hitting critical threshold.}
\label{fig:fixed-adapt}
\end{figure*}

Figure \ref{fig:fixed-adapt} shows the behavior of the architecture and the flow of iCub's comfort value for two different participants in different sessions of interaction. Figure \ref{fig:fixed-adapt}.a illustrates the behavior of the architecture for a participant that had its first interaction with the robot in the Fixed session. Here the critical threshold was hit first two times while the participant was performing the secondary task, and the participant ignored the robot's attempts to engage; and additional three times in the last phase of the session after the timer for the secondary task ran out, but in these three instances the participant was no longer distracted and answered iCub's calls.\\

There are two reasons why the three responded calls are so close in succession one after the other. The first reason is that iCub was in the Fixed personality, so it did not adapt to the person's reduced interaction during the secondary task. This explains why the first three (out of the five in total) threshold hits happened at an identical regular period. The last two threshold hits instead happen so close to each other since the participant responded unstably to iCub's calls, in a manner of giving brief stimuli and then turning their attention to something else, which did not provide iCub with enough stability to be comforted. Instead in the final instance when the critical threshold is hit the participant's response was a more stable one, interacting on several modalities, so as a result iCub's comfort resumed growing.\\

Figure \ref{fig:fixed-adapt}.b instead shows the interaction between iCub and another participant interacting with it again for the first time, but in the Adaptive session. This participant was a less interactive one in comparison to the participant in Figure \ref{fig:fixed-adapt}.a, but even so the total number of threshold hits was three, out of which only one was not answered. This demonstrates the effectiveness of the adaptivity of the architecture, which can be observed also in the decay slope during the secondary task. After two adaptations of the architecture the decay slope is a much slower one, allowing for iCub not to hit another critical point until very near the end of the interaction.\\

Independently of the order of the interaction sessions (AF or FA) or the phase of interaction, overall during the experiment on average people hit a threshold 1.42 times during one session, 1.79 times on average during the first session and 1.04 during the second one. \\

The absolute number of threshold hits summed for all participants was 68, out of which only 2 (3\%) were saturation hits, and all remaining ones (97\%) were critical. In these calculations the first two participants were excluded due to technical reasons rendering their number of threshold hits unusable.\\

\begin{figure}[H]
  \centering
    \includegraphics[width=0.45\textwidth]{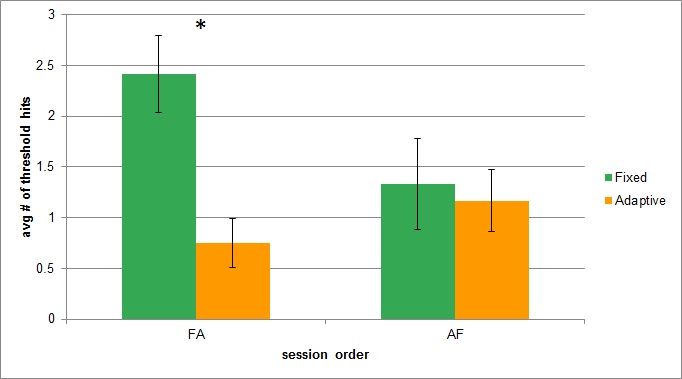}
    \caption{Comparison of average amount of threshold hits per session and order group}
    \label{fig:avgHits}
\end{figure}

Figure \ref{fig:avgHits} illustrates the effect of the order of the sessions on people's first interaction with iCub. Overall, the participants in the FA group had noticeably more threshold hits in the Fixed session than in the Adaptive, whereas the participants in the AF group had a roughly similar ratio of total threshold hits in the Fixed and Adaptive sessions.

This was additionally confirmed after running a mixed-model 2-factor ANOVA, with SESSION (levels: adaptive and fixed) and ORDER (levels: AF,FA; signifying the groups of participants) as the within and between factors respectively. A difference has been considered significant for p $<$ 0.05. \\

A significant difference was found both over the SESSIONS (F(1,22) = 7.87, p = 0.01) and for the interaction between the two sessions for the FA group (F(1,22) = 5.27, p = 0.03), confirmed with a Bonferroni test.\\

\begin{figure}[H]
    \centering
        \includegraphics[width=\textwidth]{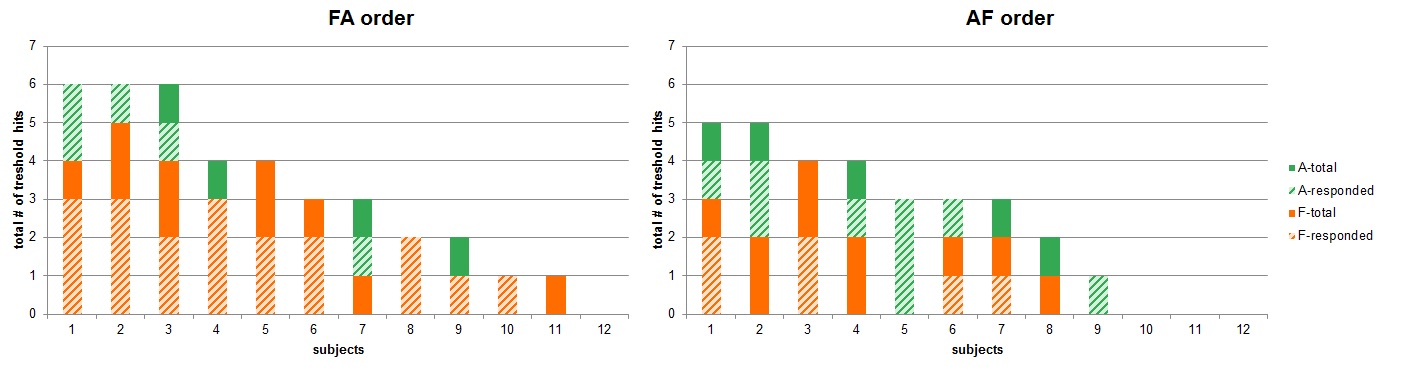}
    \caption{Number of occurred and responded threshold hits per session for FA and AF participants}
\label{fig:threshFAAF}
\end{figure}

A deeper analysis into the individual modes of behavior are presented in Figure \ref{fig:threshFAAF}. This analysis consisted of measuring the changes in the architecture for each participant, comparing for the two different orders of sessions how many times the thresholds of the architecture were hit, as well as how many times people responded to the calls for interaction in critical.\\

While a large variety in the number of threshold hits (ranging from 0 to 6) can be seen in both conditions and across both session orders, it can be noticed that the majority of people showed a tendency to respond to the robot's calls. There were some participants that never hit a critical or saturation threshold (indicated at the end of both figures), however there were only two participants which did not respond to the robot's calls for engagement, suggesting that in addition to iCub being adaptive in some cases, participants adapted always to the robot. \\

The analysis of the architecture dynamics highlighted the difference in which session was the starting session for participants as shown in \ref{fig:avgHits}: FA participants had a more challenging first session since it was both the first session of interaction with the robot, and the session where the architecture did not adapt to their interaction particularities. On the other hand, the AF participants' first session of interaction with the robot was the one where iCub was adapting its comfort variables to their interaction profiles, which contributed to them having less threshold hits in their Fixed session when compared to their FA counterparts.

\subsection{Subjective evaluation}
\label{subjEval}
The subjective evaluation included exploring the explicitly-expressed preference of participants for interacting with iCub in the A or F session, their ability to differentiate between the two different profiles of the robot, and evaluating whether their IOS/GS changed as a function of the time spent with the robot or the adaptivity of the robot.

\begin{figure*}
    \centering
        \includegraphics[width=0.9\textwidth]{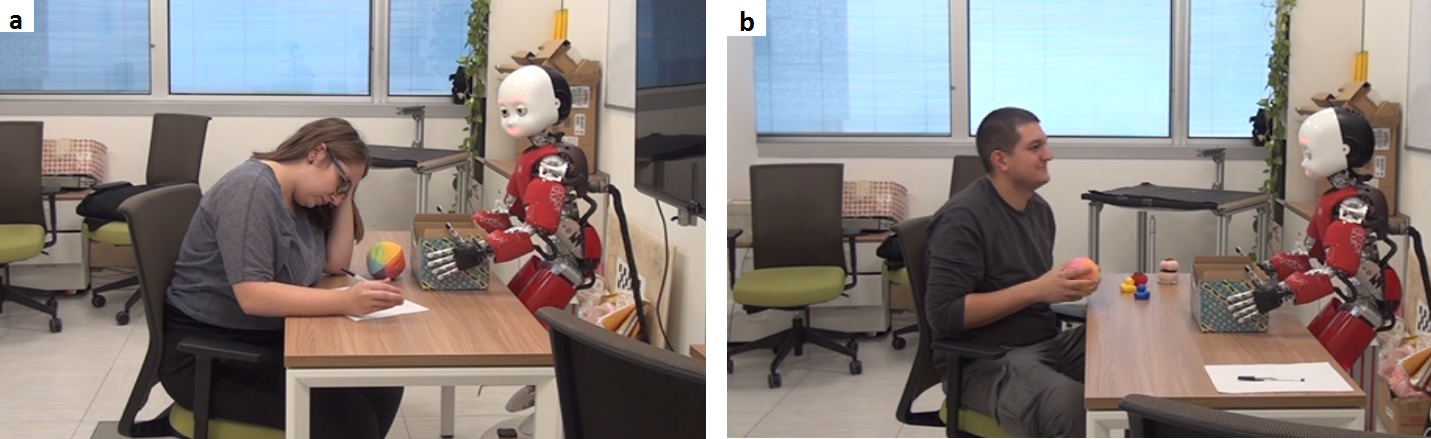}
    \caption{Participants in interaction with iCub. (a) Participant working on the pollinator puzzle. (b) Participant interacting with iCub . Informed consent of participants has been obtained for the use of their photo.}
\label{fig:participants}
\end{figure*}

In our work we wanted to explore the comparison between two similarly dynamic and responsive architectures, where the only difference between them was the inclusion of the adaptive component. \\

\begin{figure}[H]
  \centering
    \includegraphics[width=0.75\textwidth]{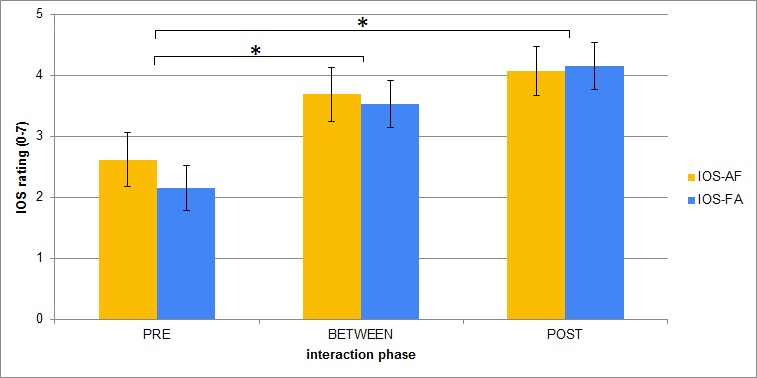}
    \caption{IOS ratings distributions across order and phases}
    \label{fig:avgIOS}
\end{figure}

\begin{figure}[H]
  \centering
    \includegraphics[width=0.75\textwidth]{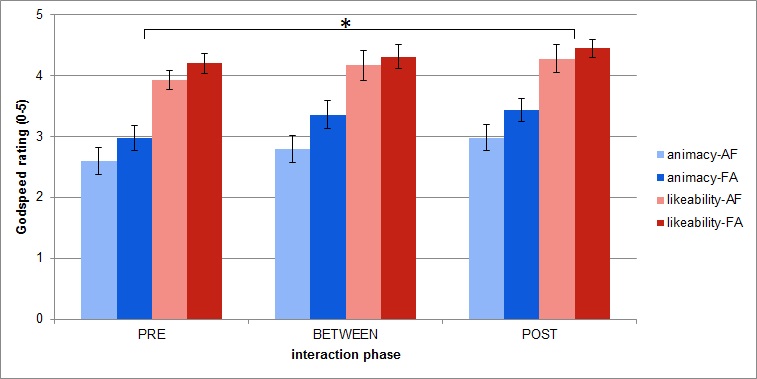}
    \caption{Godspeed ratings distributions across order and phases}
    \label{fig:avgGod}  
\end{figure}

We were curious to investigate the effect on the adaptivity level of iCub to the participants` self-rated feelings of closeness with the robot (the IOS rating) and the participants` evaluation of the robot's animacy and likeability (the Godspeed ratings). Figures \ref{fig:avgIOS} and \ref{fig:avgGod} show the ratio of the participants' IOS and Godspeed evaluations before interacting, between the two interaction sessions, and at the end of interaction. \\

Statistical analysis was performed on all IOS and Godspeed ratings. To better assess the noted difference between the ratings, a mixed-model 2-factor ANOVA was run, with PHASE (levels: pre, between and post; signifying the rating pre-experiment,between-sessions and post-experiment) and ORDER (levels: AF,FA; signifying the groups of participants)  as the within and between factors respectively. The ANOVA results follow below:

\begin{itemize}
\item The IOS rating of closeness increased significantly over the PHASE (F(2,48) = 19.88, p $<$ 0.001), while the factor ORDER (F(1,24) = 0.128, p = 0.723) was not significant, nor the interaction (F(2,48) = 0.46, p = 0.636); running a Bonferroni test found significant difference between 1st and 2nd phase and 1st and 3rd phase, but no statistically significant increase between 2nd and 3rd phase;
\item The Godspeed rating of Animacy  increased significantly over the PHASE (F(2,48) = 5.65, p = 0.006), while the factor ORDER (F(1,24) = 0.798, p = 0.38) was not significant, nor the interaction (F(2,48) = 0.03, p = 0.967); running a Bonferroni test found significant difference only between 1st and 3rd phase;
\item The Godspeed rating of Likeability increased significantly over the PHASE (F(2,48) = 6.28, p = 0.003), while the factor ORDER (F(1,24) = 2.642, p = 0.117) was not significant, nor the interaction (F(2,48) = 0.33, p = 0.72);  running a Bonferroni test found significant difference only between the 1st and 3rd phase.
\end{itemize}

From this analysis what we observed was that participants' rating of their perceived closeness with iCub changed significantly as a result of them spending more time in interaction with it, and not as a function of the adaptivity of the robot, which could signify that on their part, participants did not perceive any structural difference between the two sessions. 


This can depend on the fact that the two sessions were not particularly different to the participants who did not exploit the adaptivity of iCub excessively. Alternately, notwithstanding the differences experienced by the participants, both sessions could have been equally "likable" to them.\\
%we calculated percentage of subjects that had at least two hits, so for who the architecture changed, plotted correlation with ios/gs / we evaluaated whether a trend existed for partiicapntts that expereinced a larger difference between architecture sessions and difrence in ratings, but we couldnt find anything 
%Additionally, it can be argued that a large subset of participants did not exploit the adaptivity of the robot excessively, leading to both sessions appearing similar to the participants; or that both sessions were equally "likeable" to the participants. To further explore this, a we explored instead of correlation was done between their IOS/GS ratings and how much the architecture changed for each participant.

While people seemed more consistent in their Godspeed ratings across all sessions, their IOS ratings tended to be more variable, with bigger differences (usually of 1, but also reaching 2 and 3) between the different sessions. However also here the same conclusion was evident - the rating of IOS closeness increased for most people as a consequence of the prolonged time spent with the robot, and not as a result of the adaptability. It would seem that although there were differences in the two sessions, people did not change their rating.\\

This was confirmed also by the free questions they had to answer after the second session:

- Which session did you prefer and why?

- What was the difference (if any) you noticed between the two sessions of interaction?\\

77\% of participants answered that they preferred the second session because they felt iCub was more animated or interactive towards them, 19\% replied that they enjoyed both sessions equally and only one participant said he did not enjoy any of the two. Additionally, 27 \% answered that they did not perceive any difference between the sessions, 46\% instead had perceived the robot being more interactive in the second session (however from those 46\%  half were FA and half AF, signifying random chance), and 23\% said they learned how to interact better in the second session.

\subsection{Behavioural evaluation}
\label{behavEval}

After analyzing the subjective evaluation, the final step was processing the behavioral results, which measured how the interaction between iCub and the participants actually unfolded. The behavioral evaluation of the participants analyzed if people actually interacted differently with the robot across different phases and different modalities. This was considered again as a function of the time spent with the robot or the session order. An additional analysis was done on into how the participants' behavior changed during the dual task.\\

\begin{figure}[H]
  \centering
    \includegraphics[width=0.9\textwidth]{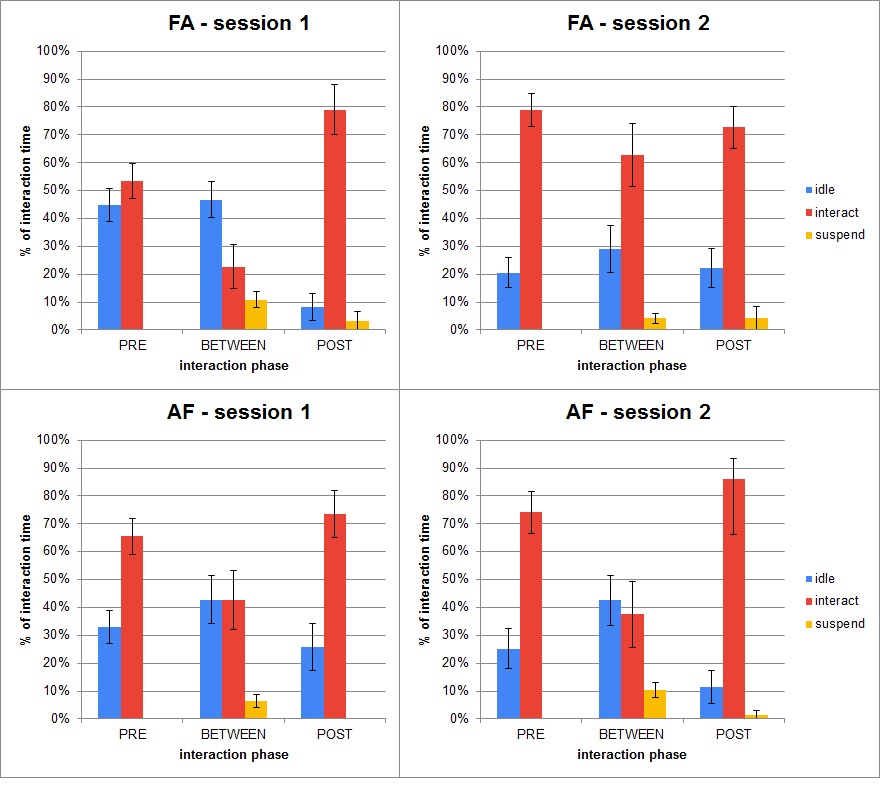}
    \caption{Distribution of the major states during the interaction, shown across phases and sessions}
    \label{fig:robotStatesFAAF}  
\end{figure}

This section covers the results from the different modalities of interaction - i.e. how people interacted with iCub on the three modalities of visual-face, visual-objects and tactile; the distribution of iCub's states during the interaction and all three phases for each session, and finally how the secondary task impacted the interaction. Figure \ref{fig:robotStatesFAAF} shows the distribution of the states the robot was in during the interaction. \\

During the interaction sessions, iCub's behaviour was guided by a state machine. The three main states were \emph{idle}, when the robot was left without stimuli from the user and interacted by itself; \emph{interact} when engagement had happened by either party; and \emph{suspend} which iCub entered after hitting a threshold hit and its call for engagement was not responded to (for critical hits). There were also more minor, transitional states signaling a change in behavior or an occurrence of the architecture adapting. However since these lasted only a few frames (and in real interaction time, less than three seconds), they were not taken into account.

From these results several conclusions can be obtained:
\begin{itemize} 
\item Participants that interacted for the first time with iCub in the Fixed condition spent less time in interaction in the very first phase of the first session when compared to the participants who had their first interaction in the Adaptive session. This effect is not present in the second session as well due to the loss of the novelty effect, since both groups of people had already interacted with the robot;
\item In the third phase (the interaction after the secondary task) there seems to be compensation for having previously ignored the robot, in the form of increased interaction. This can be seen especially in the Fixed session (regardless of order group), potentially because the robot asked for more attention without adapting to the users ignoring it;
\item The distribution pattern of the states in the last phase of the first session carry over to the first phase of the second session, indicating a training, or learning how to interact. This effect was particularly not obvious and expected for the participants of the AF group, since in their second session of interaction the architecture values of the robot were reset, so the AF participants essentially interacted first with a robot that adapted to them, and then with one that lost its adapted specifics;
\item The interactive behavior during the dual task changes significantly for the FA participants between the two sessions. Having ignored the robot during the secondary task in the first session (F) where it did not adapt to them, they seem to overcompensate in the secondary task in the second session (A) and there is a huge jump in interactivity. This may be a combined effect of both overcompensation combined with the added adaptivity of the robot;
\item The interactive behavior during the dual task stays nearly identical for the AF participants between the two sessions. Having the robot adaptive in the first session (A) it adjusted to them and it spends significantly more time in interaction than in the first session of FA participants, however due to them not perceiving the robot as particularly annoying or demanding for attention in their first session, there is not the compensation in the second session.
\end{itemize}

To evaluate the difference between the interaction patterns in the different phases, a 3-factor mixed-model ANOVA was run, with SESSION (levels: adaptive or fixed) and PHASE (levels: 1, 2 and 3; signifying the 1st, 2nd and 3rd phase of an interaction session) as the two within factors, and ORDER (levels: AF, FA) as the between factor. The percentage of time the robot spent in the interaction phase varied significantly both over the SESSION (F(1,22) = 6.08, p = 0.02) and PHASE (F(2,44) = 20.14, p $<$ 0.001), while the factor ORDER and the INTERACTION were not significant. The Bonferroni test showed significant differences between the 1st and 2nd, and the 2nd and 3rd phases, but no significant difference between 1st and 3rd phase.\\

\begin{figure}[H]
  \centering
    \includegraphics[width=0.75\textwidth]{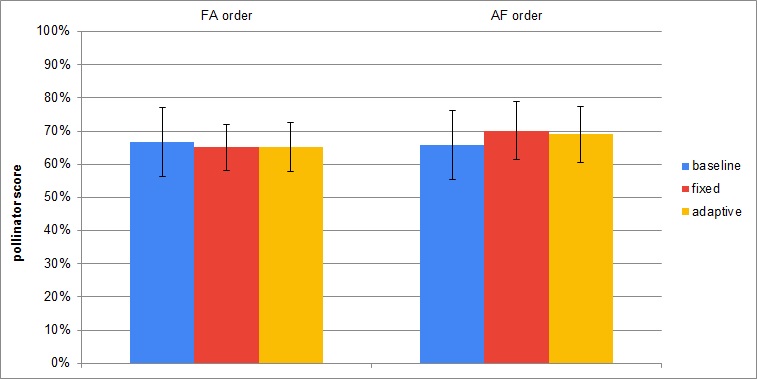}
    \caption{Average pollinator scores for the three times participants did the puzzle}
    \label{fig:pollinatorScores}  
\end{figure}

Since there was a noticeable difference in how people behaved with the robot while they were tasked with the pollinator puzzle, the next analysis focused on looking into the score of the pollinator puzzle. From Figure \ref{fig:pollinatorScores} showing the averaged pollinator scores for both groups (AF and FA) over the three times they filled the puzzle - baseline, first session, second session) it can be noticed that there is not a significant difference over the average score, signifying that even on the phases when the robot was non adaptive, on average participants could complete the task to some extent.

With this analysis it was established that the behavior of people during the secondary task (interacting with the robot or ignoring it in order to focus on the task) did not strongly impact their pollinator score. In other words, how good people were at the task was something subjective for each person themselves, and did not depend on whether they interacted a lot with the robot or ignored it. The last step of the analysis looked into the modalities participants used when interacting with iCub.\\

\begin{figure}[H]
  \centering
    \includegraphics[width=0.9\textwidth]{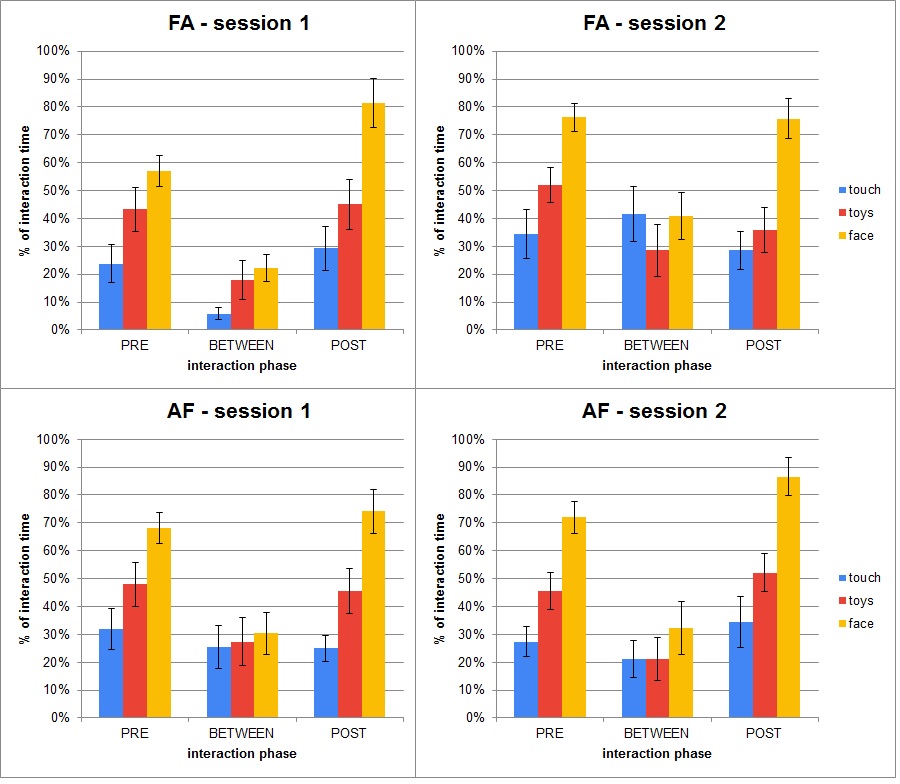}
    \caption{Distribution of the perceived stimuli in different modalities during the interaction, shown across phases and sessions}
    \label{fig:robotModalitiesFAAF}  
\end{figure}

What can be observed from the modalities graphs shown in Figure \ref{fig:robotModalitiesFAAF}  is that during the secondary task there is understandably the biggest drop in face as input, but compensation with touch, which stays similar and does not have such a significant drop. The patterns in the last phase of session 1 tend to be nearly identical to the first phase of session 2, the reason behind which can be that the mode of interaction carries over between the two sessions. a similar pattern can be also observed in the analysis of the states distribution. To evaluate the difference between the interaction patterns in the different phases, three 3-factor mixed-model ANOVA were run, with SESSION (levels: adaptive or fixed) and PHASE (levels: 1, 2 and 3; signifying the 1st, 2nd and 3rd phase of an interaction session)  as the two within factors, and ORDER (levels: AF, FA) as the between factor. A difference has been considered significant for p $<$ 0.05. 

\begin{itemize}
\item Touch: The percentage of time the robot spent in the interaction phase varied significantly only over the interaction (session*phase*order) (F(2,44) =6.22, p = 0.004), while SESSION (F(1,22) = 1.92, p = 0.18) and PHASE (F(2,44) = 0.69, p = 0.5), were not significant, neither was ORDER.  The Bonferroni test showed significant differences for the FA participants in phase 2 between session 1 and 2 (meaning between the two sessions for the FA group during the secondary task);
\item Objects: The percentage of time the robot spent in the interaction phase varied significantly  PHASE (F(2,44) = 24.46, p $<$ 0.001), while both the SESSION (F(1,22) =0.08, p = 0.78) and the interaction (F(2,44) = 0.52, p = 0.93) were not significant, and neither was ORDER. The Bonferroni test for phase showed difference between the 1st and 2nd phase, and the 2nd and 3rd phase; 
\item Face: The percentage of time the robot spent in the interaction phase varied significantly both over the SESSION (F(1,22) = 5.3, p = 0.03) and PHASE (F(2,44) =46.28, p $<$ 0.001), while the interaction (F(2,44) = 0.28, p = 0.76) was not significant, and neither was ORDER. The Bonferroni test for phase showed difference between the 1st and 2nd phase, and the 2nd and 3rd phase.
\end{itemize} 

Even though the subjective evaluation of the participants did not express a correlation between the adaptiveness of the robot with its likability, nor an awareness of the participants for there existing a difference in the profiles at all, there were implicit results pointing to the opposite. The manner of interacting with the robot, both in terms of frequency and of used modalities, changed noticeably, particularly when participants were tasked with the secondary task. More precisely, when the robot was in its adaptive profile, even if the people were given another task to complete, they still managed to interact with the robot in parallel.

%%%%%%%%%%%%%%%%%%%%%%%%%%%%%%%%%%%%%%%%%
\section{Discussion}
\label{sect:discussion}
Different individuals have different inclinations to interact with others, which can be seen also in their approach to interaction with robots. At the same time, different tasks might require different level of human intervention (or robot request for help). Creating a unique robot behavior (or personality) able to fit with task constraints and at the same time with individual desires is an impossible challenge. Endowing the robot with a possibility to adapt to its partners' preferences is therefore important to grant a certain degree of compliance with individual inclinations.\\

Our study wanted to tackle this issue by developing a personalized adaptive robot architecture. This architecture enabled the robot to adjust its behavior to suit different interaction profiles, using its internal motivation which guided the robot to engage and disengage from interaction accordingly, while also taking in account the behavior of the person interacting with it. \\

The caretaker study brought to light two different and opposing, but valuable findings. Participants were not consciously, or at least on an affective level, aware of experiencing two different robotic profiles. When asked explicitly for a difference between the two sessions of interactions, the majority of participants did not report one, or they reported their feeling that the second session had the more interactive robot profile. This however was strongly influenced by nearly all participants having reported they preferred the second session of interaction, signifying that it was not the profile of the robot that influenced their feeling, but rather the gained knowledge on how to better interact with it and the prolonged time spent in interaction. However their manner of interacting with the robot showed noticeable changes depending on the phase and session they were in, as well as depending on the robot behavior during the secondary task.\\

This has several implications, especially when designing different HRI scenarios. While this study addressed free-form interaction and how an adaptive robot would personalize to its caretaker; if imagining to port this architecture to an HRI study when the robot would need to learn by processing informations from visual or tactile stimuli, the implications from this study's findings show that the robot would be still capable to receive and process the necessary information from the person, even if the person would not be highly responsive or present at all times.\\

Additionally, the element of adaptability and personalization in the cognitive framework was not shown to bring any uncertainty and unpredictability. While on a conscious level they remained unaware, the adaptability of the robot still impacted the efficacy of the participants' interaction. Moreover, the presence of the critical and saturation thresholds promise an another level of richness that could be added to the interaction. \\

A robot that has a critical boundary can actively try to initiate interaction with the person, which could be useful not only in scenarios where a person might lose track of the robot or get distracted, but also in scenarios where a person might be very interested to interact with the robot but their shyness would prevent them from attempting to engage the robot first. \\

Complementary to that, a saturation boundary is not only useful for evaluating how much a person is interested in restarting an interrupted interaction, but can be also a crucial element in multi-person HRI scenarios, or if the robot needs to also accomplish some other task in addition to interacting with the people. The saturation threshold in particular was something that did not get used in its full potential in our study, which is probably due to the above-mentioned effects not carrying over to an 1-on-1 HRI scenario. A limitation of our study can be found in that even though the interaction was designed to be as most free-form as possible, it was still a very simplified scenario of interaction. This was also due to the limitations of current state-of-the-art: artificial cognitive agents (such as robots) are not yet at the level of replicating the human cognitive abilities, and the aspect where this was felt the most was in the absence of a verbal interaction. \\

However, adaptivity is a very important building block of cognitive interaction, and in that way endowing with it a humanoid robot like iCub, even in a scenario with behavior of lower cognitive intelligence, is still already a first step towards approaching personalized and cognitive human-robot interaction. Indeed, this effect can be seen even in children - we observe their limited capabilities (e.g. before 2 years of age they are not speech-proficient), but still they are cognitive agents that are very efficient at establishing adaptive interaction as a function of their partner, be it a peer or a caregiver. The hope and future direction of this research is that by investigating other cognitive functionalities to implement and other scenarios of interaction, the adaptive framework will reach the point of a more individualized, long-term, generalized interaction between humans and robots.

%%%%%%%%%%%%%%%%%%%%%%%%%%%%%%%%%%%%%%%%%
\section*{Conflict of Interest Statement}
%All financial, commercial or other relationships that might be perceived by the academic community as representing a potential conflict of interest must be disclosed. If no such relationship exists, authors will be asked to confirm the following statement: 
The authors declare that the research was conducted in the absence of any commercial or financial relationships that could be construed as a potential conflict of interest.

\section*{Ethics Statement}
Participants signed an informed consent form approved by the ethical committee of Liguria region (Comitato Etico Regione Liguria-Sezione 1), informing them that their performance could be recorded using cameras and microphones, as well as requesting their consent for the usage of the data for scientific purposes. All but three participants received a compensation of 10 euros and all followed the same experimental procedure.

\section*{Author Contributions}
All authors contributed to the design of the experiment. AT cured the data collection. AS and AT cured the data analysis. All authors contributed to the writing and revision of the manuscript.

\section*{Funding}
This work has been supported by a Starting Grant from the European Research Council (ERC) under the European Union's Horizon 2020 research and innovation programme. G.A. No 804388, wHiSPER

\section*{Acknowledgment}
The authors would like to thank those who participated in the study, as well as Matthew Lewis and Imran Khan from the University of Hertfordshire for their help and availability. 

\bibliographystyle{spbasicUnsort}      % basic style, author-year citations
\bibliography{biblioCogHRI}

\end{document}